%% file: ICIP.tex
\documentclass{article}
\usepackage{spconf,amsmath,graphicx}
\usepackage{xcolor}
\usepackage{graphicx}
\usepackage{booktabs}
\usepackage{tabularx}
\usepackage{array}
\usepackage{multirow}
\usepackage{svg,microtype}
\usepackage[most]{tcolorbox}
\usepackage{cleveref}
\usepackage[table]{xcolor}

\definecolor{lightgray}{gray}{0.93}
\newcolumntype{C}{>{\centering\arraybackslash}X}

\title{Is SAM3 ready for pathology segmentation?}
\name{Author(s) Name(s)}
\address{Author Affiliation(s)}

\name{Qiuyu Kong$^{\star \dagger}$ \qquad Shakiba Sharifi$^{\dagger}$ \qquad Yiming Wang$^{\ddagger}$ \qquad Marco Cristani$^{\dagger}$ \qquad Zanxi Ruan$^{\dagger}$}

\address{$^{\star}$ Sapienza University of Rome \\
 $^{\dagger}$ University of Verona \\ 
 $^{\ddagger}$ Fondazione Bruno Kessler }

\begin{document}
%
\input{preamble}
\maketitle
\input{section/abstract}

\input{section/intro}
\input{section/relatedwork}

\input{section/method}
\input{section/experiments}
\input{section/conclusion}

\bibliographystyle{IEEEbib}
\bibliography{strings,refs}

\end{document}

%% file: preamble.tex
\newcommand{\zanxi}[1]{{\color[HTML]{3CB371}[ZA: #1]}}
\newcommand{\yiming}[1]{{\color[HTML]{2F13A1}[YW: #1]}}
\newcommand{\qiuyu}[1]{{\color{green}[QI: #1]}}
\newcommand{\TODO}[1]{\textbf{\color{red}[TODO: #1]}}
\newcommand{\mc}[1]{{\color{purple}[MC: #1]}}

\definecolor{lightgreenbg}{HTML}{cde5b8}
\definecolor{lightbluebg}{HTML}{acc9e4}
\definecolor{lightpinkbg}{HTML}{edc7c6}
\definecolor{llmbg}{HTML}{F7FAF7} 
\definecolor{llmline}{HTML}{D7E2D7} 
\raggedbottom

\newcommand{\inlineColorbox}[2]{\begingroup\setlength{\fboxsep}{1pt}\colorbox{#1}{\hspace*{2pt}\vphantom{Ay}#2\hspace*{2pt}}\endgroup}

%% file: section/abstract.tex
\begin{abstract}
Is Segment Anything Model 3 (SAM3) capable in segmenting Any Pathology Images? Digital pathology segmentation spans tissue-level and nuclei-level scales, where traditional methods often suffer from high annotation costs and poor generalization. SAM3 introduces Promptable Concept Segmentation, offering a potential automated interface via text prompts. With this work, we propose a systematic evaluation protocol to explore the capability space of SAM3 in a structured manner.
Specifically, we evaluate SAM3 under different supervision settings including zero-shot, few-shot, and supervised with varying prompting strategies. Our extensive evaluation on pathological datasets including NuInsSeg, PanNuke and GlaS, reveals that: 1) text-only prompts poorly activate nuclear concepts; 2) performance is highly sensitive to visual prompt types and budgets; 3) few-shot learning offers gains, but SAM3 lacks robustness against visual prompt noise; and 4) a significant gap persists between prompt-based usage and task-trained adapter-based reference. Our study delineates SAM3's boundaries in pathology image segmentation and provides practical guidance on the necessity of pathology domain adaptation.
\end{abstract}
\addtocounter{footnote}{1}
\footnotetext{
All prompts and full implementation details will be publicly released.
}
\footnotetext[2]{Zanxi Ruan is the corresponding author.}

%% file: section/intro.tex
\section{Introduction}

Digital pathology images contain rich morphological information, ranging from \textit{macro tissue-level structures} to \textit{micro nuclei-level entities}~\cite{histopathologyreview}.
Precise segmentation of these macro and micro histological entities is a fundamental task in computational pathology. 
In practice, whole-slide and region-level analysis is typically carried out by tiling the slide into patch-level inputs, making patch-level segmentation a basic building block for learning-based computational pathology pipelines.
Existing methods, whether targeting tissue or nuclei, rely heavily on dense annotations and case-specific training. However, they often generalize poorly to different staining methods and diagnostic approaches between different institutions, while making nucleus-level labeling prohibitively costly.

\begin{figure}[t]
    \centering
    \includegraphics[width=0.9\columnwidth]{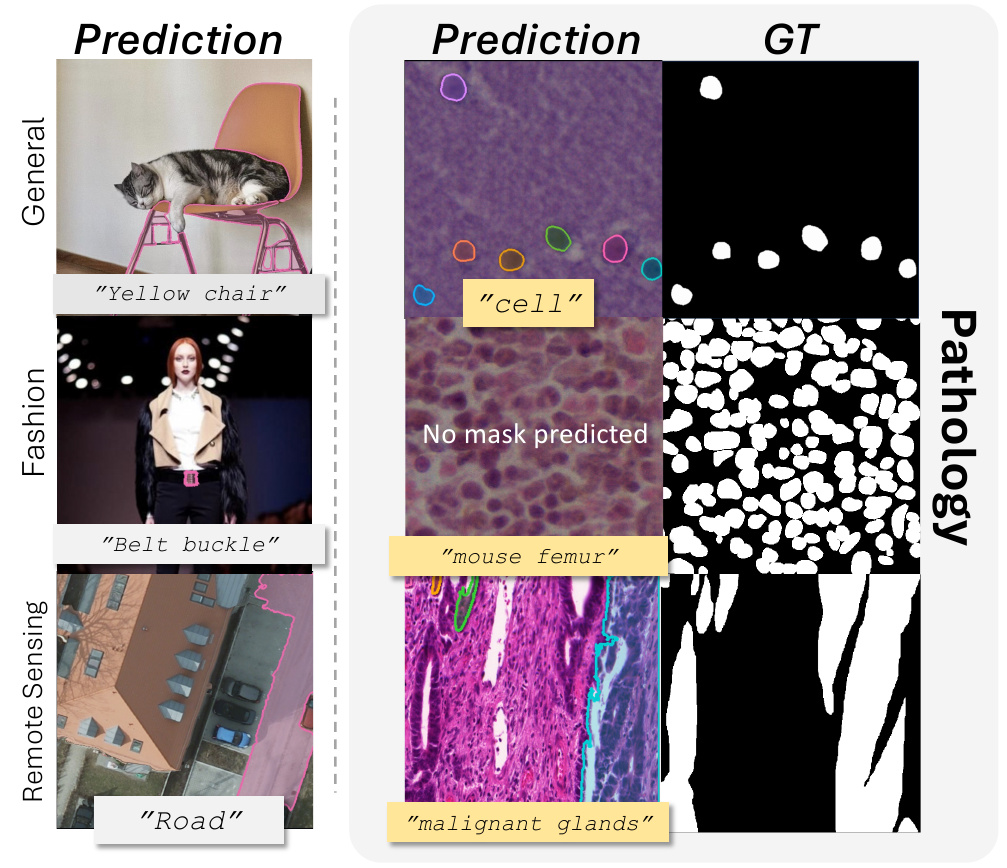}
    \vspace{-10pt}
    \caption{SAM3 with different Prompt Concepts on natural images of different domains (left) and medical images in histopathology (right). On natural images, SAM3 produces coherent masks for diverse concepts cross domains. On histopathological images, interestingly, generic prompts such as ``cell'' can segment nuclei-like structures rather well. Yet, we observe two failure modes: \textbf{i) activation failure}: specialized medical concepts, \textit{e.g.} ``mouse femur'' completely fail to produce any masks; \textbf{ii) reliability failure}: some specialized concepts activate incomplete masks, \textit{e.g.} ``malignant glands''. 
    }
    \label{fig:teaser}
\end{figure}

A general-purpose pathological segmentation system is desired to enable large-scale deployments. The recent Segment Anything (SAM)-family visual foundation models offer a new opportunity towards such general-purpose systems thanks to their large-scale pretraining. SAM family has demonstrated strong cross-task transfer via a unified prompting interface, reducing reliance on task-specific training and dense annotations~\cite{kirillov2023segment,ravi2024sam,carion2025sam}. 
Specifically, SAM1 and SAM2 supports visual prompting via points/boxes. In nuclei segmentation, where targets are small and densely distributed, such visual prompting can lead to high interaction costs. Meanwhile, the models lack domain-specific pathology knowledge, thus struggling to produce accurate and medically meaningful masks. Recently, SAM3 introduces Promptable Concept Segmentation (PCS), which enables language-based semantic segmentation, where users can segment all instances of a target category directly through text prompts. This new feature makes SAM3 an appealing candidate as a general-purpose segmentation method. 

Pathology images involve tiny objects, dense instances, and specialized semantics that differ markedly from the visual and semantic statistics of natural images used to pretrain SAM3.
As shown in Fig.~\ref{fig:teaser}, although SAM3 follows text prompts well in the general domain, the same text-driven feature behaves very differently in histopathology. 
A generic biomedical prompt (``cell'') can activate nuclei segmentation but pathology-specific term fail to trigger any prediction. At the tissue scale, even when a specific concept is activated, the resulting masks can remain incomplete, and mask edges are not sufficiently precise.
Together, these observations indicate that SAM3’s text-driven PCS are unreliable in histopathology. These considerations raise our central scientific question: \textit{Is the semantic perception capability of SAM3 sufficient to bridge the domain gap in histopathology?} 

To investigate this question, we conduct a systematic evaluation on both nuclei- and tissue-centric pathology benchmarks under controlled prompting and supervision settings, to delineate when SAM3 can be used in practice and where it breaks.
Specifically, we first test \textit{zero-shot text prompting} with SAM3 to assess the core promise of PCS by examining whether language alone can specify pathology targets without any parameter updates. Second, we benchmark \textit{zero-shot visual prompting} across SAM1/2/3 with points and boxes to quantify how much spatial guidance can compensate for the histopathology domain shift. Third, we introduce a \textit{few-shot but training-free} regime where visual prompts are derived from a small annotated support set, assessing whether support-set context alone can provide usable spatial prompts in an interaction-free setting. Finally, we evaluate a \textit{fully supervised} baseline using the currently available open-source \textit{SAM3-Adapter}~\cite{SAM3-Adapter}, providing a practical in-domain reference and contextualizing the remaining gap to prompting-based usage.

Our evaluation spans three widely used histopathology datasets: \textit{NuInsSeg}~\cite{mahbod2024nuinsseg} for single-nucleus segmentation, \textit{PanNuke}~\cite{gamper2019pannuke} for multi-class nuclei segmentation and \textit{GlaS}~\cite{sirinukunwattana2017gland} for tissue-level segmentation. Across these benchmarks, we observe that SAM3’s text prompting is fragile in histopathology: specialized medical terminology often fails to yield reliable masks, and even generic prompts can be inconsistent across datasets.
In contrast, spatial guidance via visual prompts provides a stronger performance lever, with accuracy improving as the prompt budget increases and when higher-quality localization cues are available.
Nevertheless, even under strong prompting, a substantial gap remains to supervised in-domain adaptation and to pathology-specific segmentation methods, indicating that SAM3 is not yet a drop-in solution for pathology segmentation.
Our structured assessment of SAM3 for pathology segmentation reveals how supervision strength and prompt design affect performance. These findings offer practical guidance for choosing prompting strategies and supervision levels in pathology segmentation workflows.

%% file: section/relatedwork.tex
\section{Related Works}

In the field of pathology segmentation, fully supervised learning remains the prevailing paradigm, leveraging dense pixel-wise annotations to capture intricate morphological representations~\cite{jehanzaib2025robust}. However, acquiring large-scale and consistent expert annotations is costly, motivating increasing interest in few-shot learning for pathology segmentation. With only a few labeled support images, few-shot methods~\cite{ming2025few} aim to transfer to novel structures, enabling more practical cross-scenario generalization. Meanwhile, foundation models like SAM family~\cite{kirillov2023segment,ravi2024sam} has established a prompt-driven paradigm for high-quality, class-agnostic mask prediction. However, directly applying SAM to histopathology remains challenging: Deng et al.~\cite{deng2025segment} revealed that while SAM excels at large connected regions (organs), it struggles with dense nuclei. To bridge this domain gap, existing efforts in digital pathology can be broadly categorized into: \textit{i)} complete fine-tuning on domain-specific data, as exemplified by Med-SAM \cite{MedSAM2024}; \textit{ii)} parameter-efficient adaptation via lightweight modules such as LoRA/adapters to learn new knowledge while minimizing computational cost, \textit{e.g.}, SAM-Adapter~\cite{SAM3-Adapter}; \textit{iii)} Semantic Hybrids, which fuse SAM’s mask generation with Vision-Language Models to inject semantic knowledge, \textit{e.g.}, OV-SAM \cite{OVSAM2024}.

Recently, SAM3~\cite{carion2025sam} extends the SAM family with a more advanced architecture and enhanced generalization capability. In particular, its support for both visual and textual prompts enables more flexible and powerful cross-modal interaction, substantially strengthening its zero-shot potential. However, its specific adaptation and effectiveness within the complex domain of computational pathology remain an open frontier for investigation. Therefore, this paper aims to do an empirical study on applying SAM3 to pathology segmentation.

%% file: section/method.tex
\section{Evaluation}
\begin{figure*}[t]
    \centering
    \includegraphics[width=1.0\linewidth]{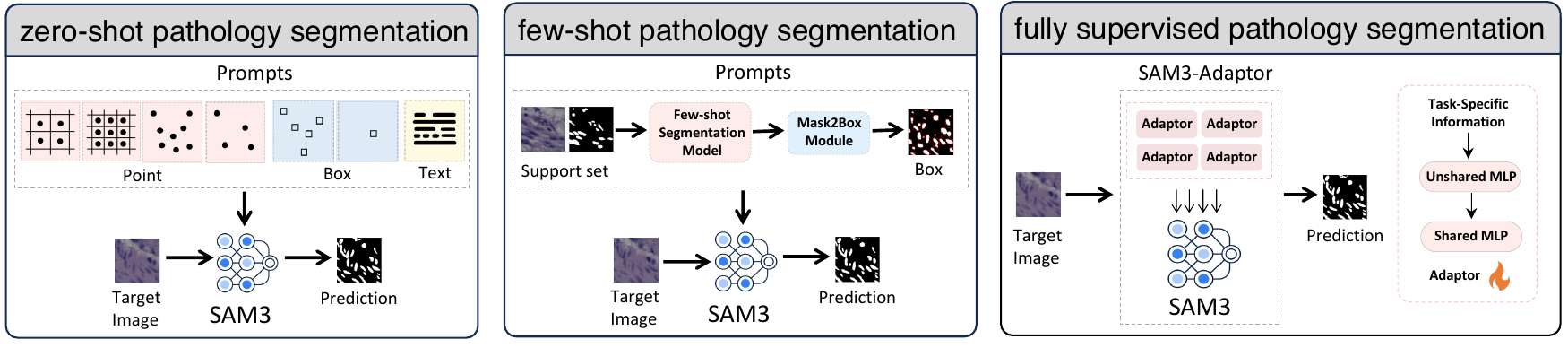}
                \caption{
                We investigate frozen SAM3 for pathology segmentation across zero-shot, supervised, and 1-shot few-shot settings (using connectivity-based Mask2Box), analyzing various prompt types and strategies.
                }
                
    \label{fig:setting}
    \vspace{-15pt}
\end{figure*}

For the evaluation, we do not modify SAM3 in terms of its model architecture, prompting interaction, and decoding strategy. 
All experiments follow the standard inference pipeline of SAM3. 
Given an image $I_q$ and a prompt $P$, SAM3 produces a segmentation mask:
\begin{equation}
    \hat{M} = f_{\text{SAM3}}(I_q, P).
\end{equation}
As shown in Fig \ref{fig:setting}, different evaluation only varies in terms of how prompts are constructed and whether domain-specific supervision is introduced. 


\noindent\textbf{\textit{Zero-shot setting:}} 
We keep SAM3 frozen and segment a target image using prompts provided on that same image, including text prompts (\textit{i.e.} promptable concepts) and visual prompts (\textit{i.e.} point/box) with controlled budgets.

\noindent\textbf{\textit{Few-shot setting:}} We aim to assess whether limited support-derived prompts can effectively guide SAM3 under low-supervision settings. Specifically, we keep SAM3 frozen and generate prompts from a small set of annotated support images rather than oracle information from the query image. 

\noindent\textbf{\textit{Supervised setting:}} 
We seek to quantify the gap between prompt-based usage and task-specific performance trained with the full train set of the dataset.
To this end, we adapt SAM3 with lightweight adapter modules to obtain a fully supervised baseline.

%% file: section/experiments.tex
\section{Experiments}
\subsection{Experimental setup}
\noindent\textbf{Datasets.} We evaluate SAM3 on three representative histopathology benchmarks covering nuclei-level and tissue-level segmentation.
NuInsSeg~\cite{mahbod2024nuinsseg} contains 665 H\&E-stained image patches (512$\times$512) with over 30,000 manually segmented nuclei from 31 human and mouse organs, and is used for single-class nuclei segmentation. PanNuke~\cite{gamper2019pannuke} contains 7901 images with 189744 labeled nuclei across 19 tissue types and five cell categories, captured at $\times$40 magnification (0.25 $\mu$m/pixel), and is used for multi-class nuclei segmentation. 
GlaS~\cite{sirinukunwattana2017gland} is a colorectal gland tissue segmentation dataset containing 165 H\&E-stained images from 16 patients with T3/T4 stage colorectal adenocarcinoma. Each image is labeled as benign/malignant and provides pixel-level instance annotations, totaling 1530 labeled objects.

\noindent \textbf{Evaluation Protocol.} We study SAM3 under three supervision settings: zero-shot, few-shot, and fully-supervised. For PanNuke, we follow the official 3-fold split. For NuInsSeg and GlaS, we perform 3-fold cross-validation by our own. Within each fold, we further split the data into training/validation/test sets in an approximately 5:1:3 ratio to ensure sufficient test samples for few-shot evaluation. In the zero-shot and few-shot settings, no parameters are updated and we evaluate directly on each fold’s test set, reporting results averaged across folds. In the supervised regime, we train the adapter on the fold-specific training set, select the best checkpoint based on the validation set, and report performance on the test set, again averaged across folds. Performance is measured by mean Intersection over Union (mIoU) and Dice coefficient. For PanNuke, mIoU is the macro average of class-wise IoU:
$\mathrm{mIoU}=\frac{1}{|\mathcal{C}|}\sum_{c\in\mathcal{C}} \frac{\mathrm{TP}_c}{\mathrm{TP}_c+\mathrm{FP}_c+\mathrm{FN}_c}$,
with $|\mathcal{C}|=5$. NuInsSeg and GlaS reports foreground IoU/Dice.

\noindent \textbf{Implementation details. } We conduct all experiments with SAM 3 as the frozen backbone model.
Specifically, in the few-shot setting, we employ a ResNet backbone pre-trained on ImageNet to extract features for support-query prototype matching~\cite{fan2022self}, which is also frozen during evaluation. In the supervised setting, lightweight adapters are trained following the SAM3-Adapter protocol, using the AdamW optimizer with a learning rate of 0.0002. All experiments are conducted on a NVIDIA 5090 GPU. The same configuration applies to SAM1/2 that are compared in our experiments.
\input{tables/table1}

\input{tables/table2}

\input{tables/table3}

\subsection{Empirical Findings and Analysis}
In preliminary experiments, we evaluated SAM3 for whole-slide image segmentation and observed that inference is prohibitively expensive, requiring 2.2 hours to process a single small WSI. So we restrict our study to a patch-level evaluation setting, which requires 1.21s per image.
This section reports our results around four questions (Q1–Q4), aiming to delineate SAM3's capability and failure boundaries in pathology image segmentation.

\begin{tcolorbox}[
  enhanced,
  colbacktitle=gray,
  coltitle=white, 
  fonttitle=\bfseries, 
  boxrule=0.5pt,
  colframe=gray,
  colback=white,
  borderline south={0.0pt}{0pt}{llmline},
  title={Question 1}, 
  drop shadow=black!30!white,
  left=2.5mm,right=2.5mm,top=1.8mm,bottom=1.8mm,
]
Is SAM3 reliable to segment nuclei and tissue-level pathology images with text prompts alone?
\vspace{-5pt}
\end{tcolorbox} 
We use text-only prompts in a zero-shot setting with no parameter updates, to test whether SAM3’s PCS can reliably work for nuclei in pathology images. To reduce randomness from prompt wording, we use three levels of semantic specificity: \textit{i)} Medical terminology: expert-/dataset-defined class names. \textit{ii)} LLM-generated vocabulary: multiple near-synonymous rewrites for each term by LLM, averaged for evaluation (we use the GPT-5.2; examples are shown in the box below.).
\noindent
\begin{tcolorbox}[
  enhanced,
  boxrule=0pt,
  colback=llmbg,
  borderline south={0.4pt}{0pt}{llmline},
  left=2.5mm,right=2.5mm,top=1.8mm,bottom=1.8mm,
]
\small
\textbf{PanNuke Dataset}\\[0.35em]
\textit{``Epithelial cell":} $\rightarrow$
{\ttfamily\footnotesize\raggedright
["Epithelial cell", "Epithelium", "Lining cell", "Surface cell", "Mucosal cell nucleus"]\par}
\end{tcolorbox} 
\noindent\textit{iii)} General medical terminology: the most general prompt ``cell" for NuInsSeg and PanNuke and ``gland" for GlaS. 

As shown in Table~\ref{tab:zero-shot-text}, when we use specialized medical terms or their LLM-based paraphrases as prompts, SAM3 almost fails on both NuInsSeg and PanNuke (mIoU \textless 10\%). This suggests that, although SAM3 shows strong zero-shot concept alignment on natural images, its pretrained semantic representations do not transfer well to pathology-specific terminology. Notably, when the prompt is simplified to a broad biomedical word, ``cell," performance on NuInsSeg jumps sharply (mIoU = 68.15\%), shows that SAM3 can segment nuclei-like structures, but it cannot reliably link nuclear visual patterns to specific pathology terms. Also, this gain does not carry over to PanNuke. Under PanNuke's multi-tissue and cluttered backgrounds, the same generic prompt quickly breaks down (mIoU = 6.22\%).
Besides, LLM-generated vocabulary generally outperforms medical terminology, yielding improvements of up to 6.43\% mIoU. Because they are more likely to activate the model's understanding of existing visual concept representations within SAM3’s fixed semantic space. 
An interesting observation is that when using the generic term “gland” on GlaS dataset, SAM3 exhibits particularly poor performance (mIoU=0.08\%).

\noindent\textbf{\textit{Observation 1.}}
We find that expanding vocabulary diversity via LLM-based prompting can improve segmentation prediction probabilities. However, the effectiveness of text prompts in SAM3 remains highly limited for histopathology images, showing poor performance on both multi-class and simple tissue-level datasets.

\begin{tcolorbox}[
  enhanced,
  colbacktitle=gray,
  coltitle=white, 
  fonttitle=\bfseries, 
  boxrule=0.5pt,
  colframe=gray,
  colback=white,
  borderline south={0.0pt}{0pt}{llmline},
  title={Question 2}, 
  drop shadow=black!30!white,
  left=2.5mm,right=2.5mm,top=1.8mm,bottom=1.8mm,
]
How much does SAM3 depend on the type and budget of visual prompts (points vs. boxes)?
\vspace{-5pt}
\end{tcolorbox} 

\begin{figure}[h]
    \vspace{-10pt}
    \centering
    \includegraphics[width=0.8\columnwidth]{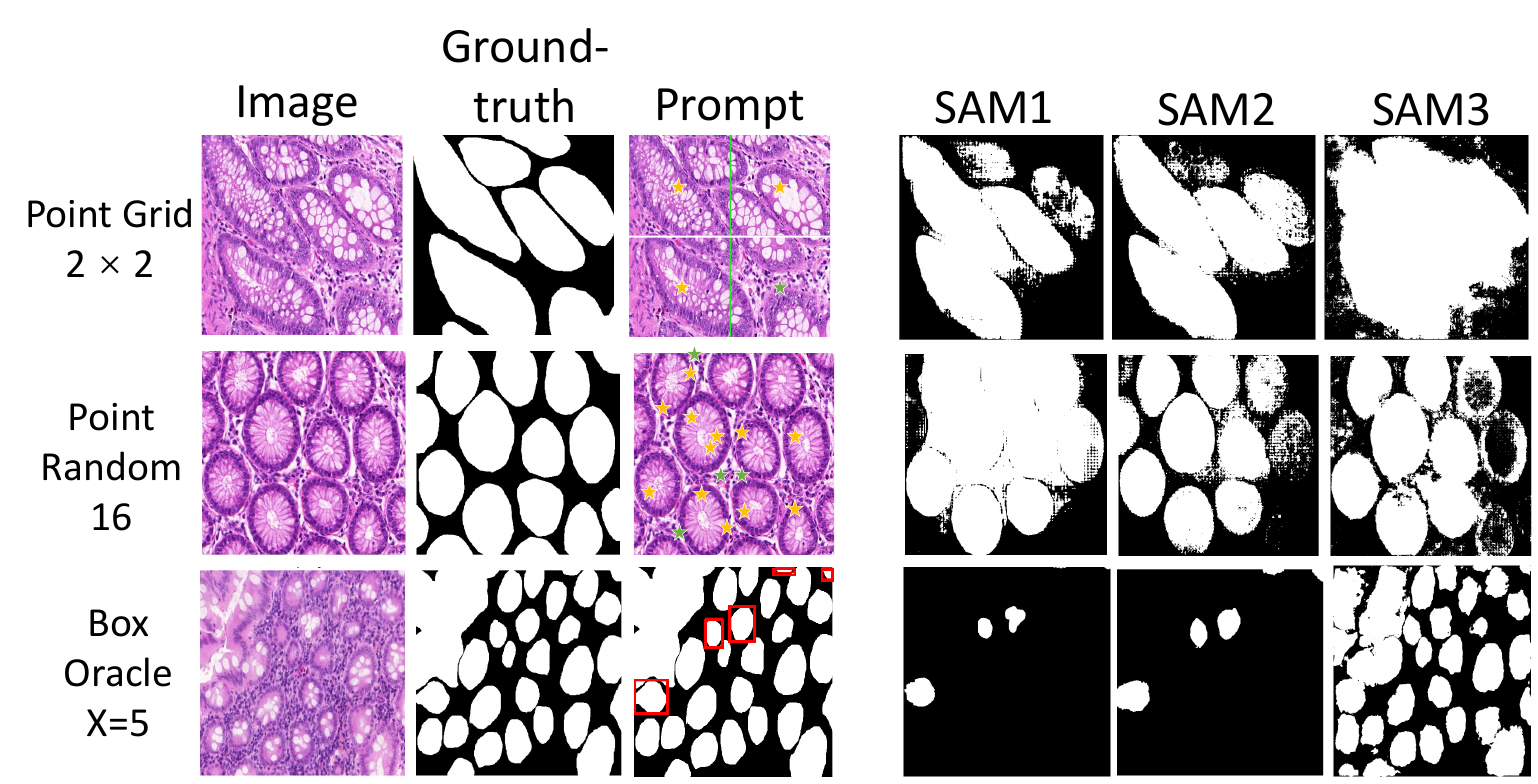}
    \vspace{-10pt}
    \caption{Zero-shot pathology segmentation on GlaS dataset.}
    \label{fig:zero_shot}
    \vspace{-15pt}
\end{figure}

Table~\ref{tab:zero-shot-visual} reports zero-shot performance on NuInsSeg, PanNuke and GlaS using visual prompts and Fig.~\ref{fig:zero_shot} presents qualitative visualizations.

\textbf{Point Prompts.}
We evaluate the impact of point budget and sampling strategy on the SAM series models: using 4/16 points respectively to represent coarse vs. denser supervision, and comparing Grid sampling (uniform distribution) with Random sampling (random over the image).
Table~\ref{tab:zero-shot-visual} shows that: \textit{i)} Point prompts perform substantially worse on nuclei-centric datasets than on tissue-level segmentation with a performance gap exceeding 35\% mIoU between the two. Because nuclei are small and discrete, making it difficult for a small number of points to cover instances, whereas structures such as glands are larger and more connected, allowing points to fall more easily inside targets and activate region expansion. \textit{ii)} Increasing the number of points consistently improves performance across all datasets for the SAM family, as a higher point budget inherently increases coverage and the likelihood of hitting target regions. \textit{iii)} Grid sampling outperforms random sampling, since random sampling has a larger variance under low point budgets, making it easy to completely miss valid targets.

\textbf{Box Prompts (Oracle).}
We further evaluate performance with box prompts by generating oracle boxes directly from ground-truth masks, thereby eliminating interference in prompt quality and enabling a cleaner assessment of the gains brought by accurate spatial prompts.
Specifically, we designed three box-count configurations: 1, X (the mode of the per-image instance count in the dataset), and all instance boxes to quantify the impact of localization information on segmentation performance.
In general, segmentation performance generally increases with the number of oracle boxes.
\noindent\textbf{\textit{Observation 2.}} Compared to text prompts, visual prompts provide a more effective way to improve the performance of SAM models in histopathology segmentation, where performance improves as the prompt budget increases and stronger prompts are provided. However, even with visual prompting, the overall performance remains inferior to task-specific methods~\cite{li2022online,mahbod2024nuinsseg}, indicating that the SAM family is not yet able to deliver stable and reliable performance in pathology segmentation scenarios.

\begin{figure}[h]
    \vspace{-10pt}
    \centering
    \includegraphics[width=\columnwidth]{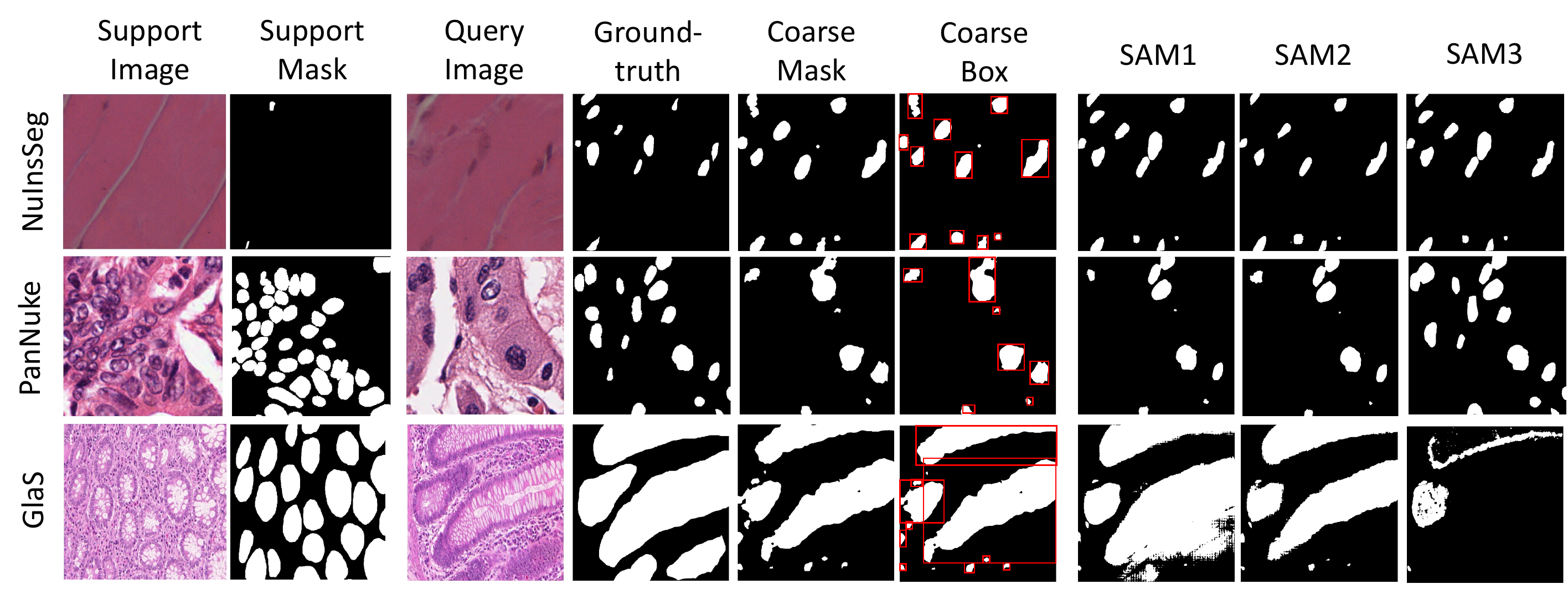}
    \vspace{-15pt}
    \caption{Qualitative few-shot pathology segmentation results.}
    \label{fig:fss}
\end{figure}

\begin{tcolorbox}[
  enhanced,
  colbacktitle=gray,
  coltitle=white, 
  fonttitle=\bfseries, 
  boxrule=0.5pt,
  colframe=gray,
  colback=white,
  borderline south={0.0pt}{0pt}{llmline},
  title={Question 3}, 
  drop shadow=black!30!white,
  left=2.5mm,right=2.5mm,top=1.8mm,bottom=1.8mm,
]
Can few-shot visual prompts effectively guide SAM3 in pathological segmentation without training?
\vspace{-5pt}
\end{tcolorbox} 
We probe SAM3 under a few-shot, training-free setting.
Motivated by Observation~2, which shows that accurate box prompts provide strong spatial guidance, we consider a practical scenario where such high-quality prompts are unavailable at test time.
Accordingly, we adopt a few-shot semantic segmentation (FSS) approach~\cite{fan2022self}, leveraging a prototype-based network to generate coarse masks and then construct boxes as visual guidance.
Fig.~\ref{fig:fss}, demonstrates that the prototype-based network is able to produce meaningful coarse masks in most cases. 
As shown in Table~\ref{tab:few-shot}, the three SAM variants show little difference within 5\% mIoU on the two nuclei-centric datasets. In contrast, SAM3 performs significantly worse on GlaS with a drop of 15.84\% mIoU compared to the best-performing method, primarily because it is more sensitive to small-scale mask noise, which further interferes with the overall segmentation of large target structures (as shown in Fig~\ref{fig:fss} row 3).
Additionally, the overall lower performance of SAM family on PanNuke can be attributed to the multi-class and complex scene characteristics of this dataset, leading to unstable box generation by the FSS model.

\noindent\textbf{\textit{Observation 3.}} In training-free scenarios, FSS can serve as a contextual prior to obtain meaningful prompts. However the performance greatly depends on the noisiness of the boxes, especially for large tissue structures.

\begin{figure}[h]
    \vspace{-10pt}
    \centering
    \includegraphics[width=0.8\columnwidth]{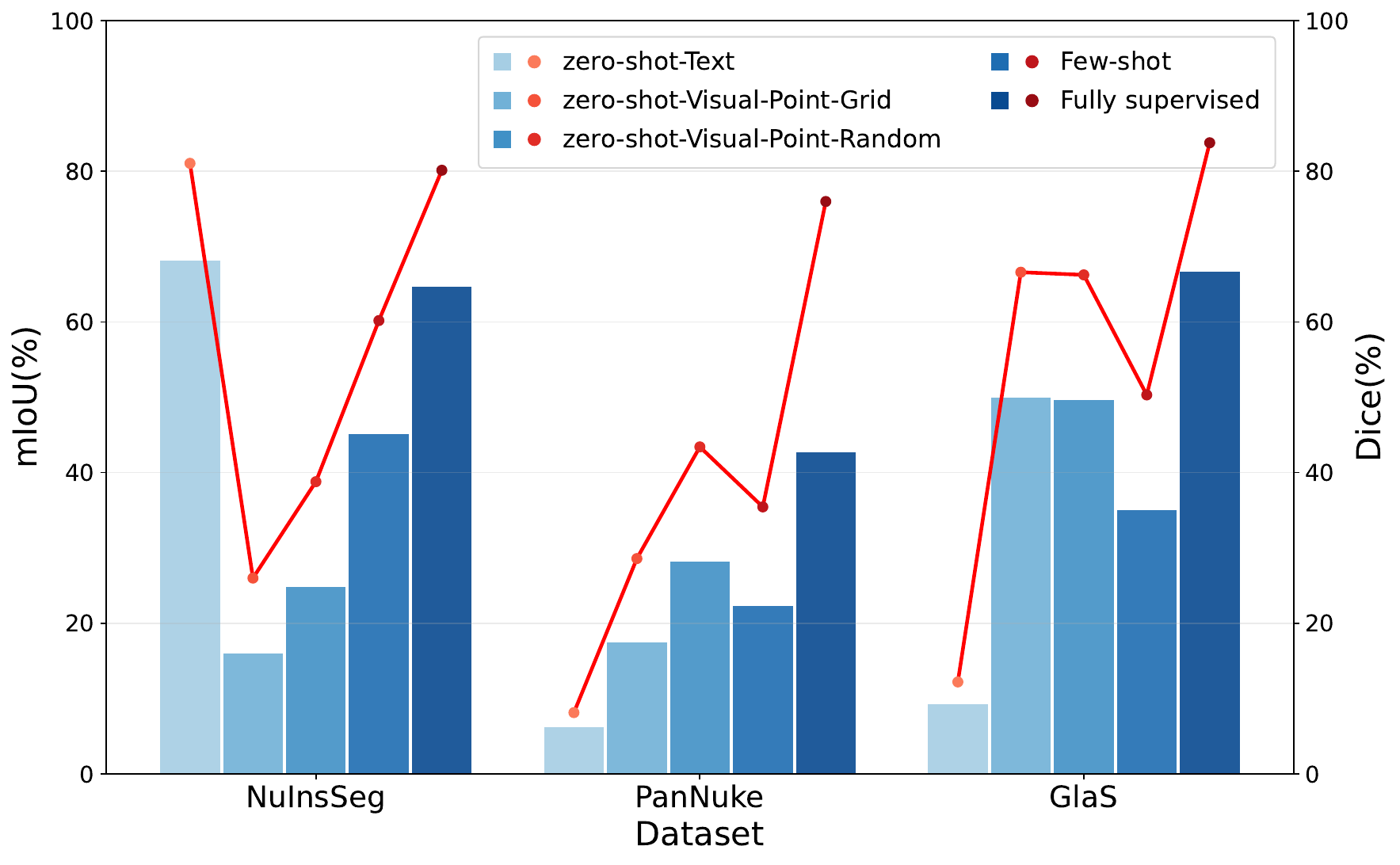}
    \vspace{-10pt}
    \caption{Performance comparison under full supervision.}
    \vspace{-15pt}
    \label{fig:fully supervised}
\end{figure}

\begin{tcolorbox}[
  enhanced,
  colbacktitle=gray,
  coltitle=white, 
  fonttitle=\bfseries, 
  boxrule=0.5pt,
  colframe=gray,
  colback=white,
  borderline south={0.0pt}{0pt}{llmline},
  title={Question 4}, 
  drop shadow=black!30!white,
  left=2.5mm,right=2.5mm,top=1.8mm,bottom=1.8mm,
]
Is it large the gap between prompting and task-specific training (adapter-based adaptation as a reference)?
\vspace{-5pt}
\end{tcolorbox} 
We evaluate a fully supervised reference by adapting SAM3 with the SAM3-Adapter~\cite{SAM3-Adapter}, noting that it adapts only the image branch and therefore does not represent a joint text-image fine-tuning upper bound.
As shown in Fig.~\ref{fig:fully supervised}, 
We found that in most cases, SAM3-Adapter achieves significantly better performance than the zero-shot and few-shot settings across the three datasets.
However, even with adaptation, SAM3-Adapter is still underperforms fully supervised histopathology-specific methods~\cite{li2022online,mahbod2024nuinsseg}, indicating that it is not yet well suited as a strong general-purpose choice for pathology datasets (NuInsSeg Dice ours:80.14\%/\cite{mahbod2024nuinsseg}81.4\%; GlaS mIoU ours:66.68\%/\cite{li2022online}86.84\%).

\noindent\textbf{\textit{Observation 4.}} 
There is a significant gap between zero-shot and fully supervised methods for pathological segmentation. Adaptation is necessary, but it remains insufficient to achieve the strength of pathology-specific methods. Thus, using SAM3 is not ready for pathological image segmentation. 

%% file: tables/table1.tex
\begin{table}[th!]
\centering
\vspace{-15pt}
\caption{Zero-shot segmentation with text prompts.}
\label{tab:zero-shot-text}
\resizebox{\columnwidth}{!}{%
\begin{tabular}{l|cc|cc|cc}
\toprule
\multirow{2}{*}{\textbf{Text Prompt}} & \multicolumn{2}{c|}{\textbf{NuInsSeg}} & \multicolumn{2}{c|}{\textbf{PanNuke}} & \multicolumn{2}{c}{\textbf{GlaS}}\\
\cmidrule(lr){2-3} \cmidrule(lr){4-5} \cmidrule(lr){6-7}
& mIoU & Dice & mIoU & Dice & mIoU & Dice \\
\midrule
Medical terminology & 3.00 & 5.79 & 0.26 & 0.37 & 4.13 & 5.28\\
LLM-generated vocabulary & 9.43 & 11.60 & 4.08 & 5.16 & 9.26 & 12.20 \\
General medical terminology & 68.15 & 81.06 & 6.22 & 8.13 & 0.08 & 0.15 \\
\bottomrule
\end{tabular}}
\vspace{-15pt}
\end{table}

%% file: tables/table2.tex
\begin{table}[th!]
\centering
\caption{Zero-shot segmentation with visual prompts (point and box). For NuInsSeg, PanNuke, and GlaS, Oracle-X uses $X=28$, $X=18$, and $X=5$ boxes, respectively.}
\label{tab:zero-shot-visual}
\resizebox{\columnwidth}{!}{%
\begin{tabular}{ll|cc|cc|cc}
\toprule
\multirow{2}{*}{\textbf{Model}} & \multirow{2}{*}{\textbf{Prompt Setting}} & \multicolumn{2}{c|}{\textbf{NuInsSeg}} & \multicolumn{2}{c|}{\textbf{PanNuke}} & \multicolumn{2}{c}{\textbf{GlaS}}\\
\cmidrule(lr){3-4} \cmidrule(lr){5-6} \cmidrule(lr){7-8}
& & mIoU & Dice & mIoU & Dice & mIoU & Dice \\
\midrule
\rowcolor{gray!15}
\multicolumn{8}{l}{\textit{Point Prompts}} \\
\midrule
SAM 1 & Grid $2\times2$ & 12.69 & 20.99 & 11.80 & 20.05 & 47.71 & 64.50 \\
SAM 2 & Grid $2\times2$ & 12.70 & 20.59 & 8.38 & 14.78 & 49.60 & 66.25\\
SAM 3 & Grid $2\times2$ & 6.52 & 10.88 & 8.48 & 14.63 & 49.54 & 66.19 \\
\addlinespace[2pt]
SAM 1 & Grid $4\times4$ & 26.94 & 40.44 & 21.78 & 34.78 & 49.76 & 66.40\\
SAM 2 & Grid $4\times4$ & 16.82 & 26.16 & 14.96 & 25.11 & 50.85 & 67.36 \\
SAM 3 & Grid $4\times4$ & 15.97 & 25.99 & 17.45 & 28.58 & 49.97 & 66.60\\
\addlinespace[2pt]
SAM 1 & Random 4 & 17.26 & 28.38 & 17.32 & 28.97 & 43.04 & 60.11\\
SAM 2 & Random 4 & 18.91 & 29.95 & 14.03 & 23.95 & 48.61 & 65.35\\
SAM 3 & Random 4 & 11.91 & 20.65 & 15.42 & 26.52 & 46.18 & 63.14 \\
\addlinespace[2pt]
SAM 1 & Random 16 & 24.83 & 37.77 & 20.02 & 32.45 & 48.19 & 65.00\\
SAM 2 & Random 16 & 20.79 & 32.27 & 16.07 & 26.82 & 50.42 & 66.98 \\
SAM 3 & Random 16 & 24.83 & 38.79 & 28.13 & 43.41 & 49.58 & 66.25\\
\midrule
\rowcolor{gray!15}
\multicolumn{8}{l}{\textit{Box Prompts (Oracle)}} \\
\midrule
SAM 1 & Oracle 1 & 5.63 & 9.44 & 24.77 & 33.13 & 12.76 & 22.48\\
SAM 2 & Oracle 1 & 5.91 & 10.01 & 26.14 & 34.60 & 13.47 & 23.58\\
SAM 3 & Oracle 1 & 6.26 & 10.33 & 27.04 & 35.24 & 38.76 & 52.47 \\
\addlinespace[2pt]
SAM 1 & Oracle X & 65.56 & 77.76 & 75.64 & 85.29 & 50.63 & 66.91\\
SAM 2 & Oracle X & 65.92 & 78.26 & 78.44 & 87.18 & 52.55 & 68.58\\
SAM 3 & Oracle X & 69.16 & 80.45 & 82.37 & 89.68 & 65.50 & 78.79 \\
\addlinespace[2pt]
SAM 1 & Oracle all & 77.08 & 86.93 & 79.43 & 88.25 & 77.20 & 87.03\\
SAM 2 & Oracle all & 76.61 & 86.57 & 82.22 & 90.02 & 80.81 & 89.28 \\
SAM 3 & Oracle all & 81.19 & 89.54 & 86.52 & 92.69 & 68.33 & 80.87 \\
\bottomrule
\end{tabular}}
\vspace{-15pt}
\end{table}

%% file: tables/table3.tex
\begin{table}[th!]
\footnotesize
\centering
\caption{Few-shot segmentation with box prompts.}
\label{tab:few-shot}
\resizebox{0.8\columnwidth}{!}{%
\begin{tabular}{l|cc|cc|cc}
\toprule
\multirow{2}{*}{\textbf{Model}} & \multicolumn{2}{c|}{\textbf{NuInsSeg}} & \multicolumn{2}{c|}{\textbf{PanNuke}} & \multicolumn{2}{c}{\textbf{GlaS}}\\
\cmidrule(lr){2-3} \cmidrule(lr){4-5} \cmidrule(lr){6-7}
& mIoU & Dice & mIoU & Dice & mIoU & Dice \\
\midrule
SAM 1 & 45.26 & 61.19 & 18.04 & 30.26 & 49.76 & 66.06\\
SAM 2 & 43.38 & 59.48 & 17.56 & 29.57 & 50.84 & 67.13\\
SAM 3 & 45.06 & 60.18 & 22.32 & 35.44 & 35.00 & 50.31\\
\bottomrule
\end{tabular}}
\vspace{-15pt}
\end{table}

%% file: section/conclusion.tex
\section{Conclusion}
We systematically evaluate SAM3 on nuclei- and tissue-level histopathology benchmarks under text and visual prompting. We find that text prompting is unreliable in pathology, and while stronger visual prompts  substantially improve performance, a clear gap remains to supervised SAM3 adapter-based adaptation. Moreover, even after adaptation, SAM3 still lags behind pathology-specific fully supervised methods, indicating that it is not yet a drop-in solution for pathology segmentation.
Practically, we recommend prioritizing box prompts over point prompts and increasing box quality/budget whenever possible; for text prompting, expanding prompt vocabulary with LLM-generated variants can provide modest robustness. When training is not feasible, deriving prompts from a few-shot segmentation prior offers a stable training-free alternative. 
Nevertheless, to fully unlock the potential of SAM3, domain-specific fine-tuning remains essential.